\documentclass[12pt, a4paper, logo, copyright, nonumbering]{xds}

\usepackage{subcaption}
\usepackage[sort]{natbib}
\usepackage{dblfloatfix}
\usepackage{ulem}
\usepackage{float}
\usepackage{caption}
\usepackage{dramatist}
\usepackage{xspace}
\usepackage{pifont} 
\usepackage{multirow}
\usepackage{tcolorbox}
\usepackage{xltabular}
\usepackage{longtable}
\usepackage{mdframed}
\usepackage{hyperref}
\usepackage{tocloft}
\usepackage{amsfonts}
\usepackage{amsmath}
\usepackage{amssymb}
\usepackage{lineno}
\usepackage{multirow}
\usepackage{adjustbox}

\usepackage[bottom]{footmisc}

\usepackage{CJKutf8}
\usepackage{setspace}

\usepackage{graphicx}

\usepackage{dsfont}
\usepackage{array} 
\usepackage{tabularx} 
\usepackage{xcolor} 

\usepackage{lipsum}  
\usepackage{multicol} 

\makeatletter
\def\@BTrule[#1]{%
  \ifx\longtable\undefined
    \let\@BTswitch\@BTnormal
  \else\ifx\hline\LT@hline
    \nobreak
    \let\@BTswitch\@BLTrule
  \else
     \let\@BTswitch\@BTnormal
  \fi\fi
  \global\@thisrulewidth=#1\relax
  \ifnum\@thisruleclass=\tw@\vskip\@aboverulesep\else
  \ifnum\@lastruleclass=\z@\vskip\@aboverulesep\else
  \ifnum\@lastruleclass=\@ne\vskip\doublerulesep\fi\fi\fi
  \@BTswitch}
\makeatother

\addto\extrasenglish{
}

 {\begin{list}{}%
         {\setlength{\leftmargin}{#1}}%
         \item[]%
 }
 {\end{list}}

\reportnumber{001} 
\renewcommand{\today}{}

\usepackage{tikz}

\usepackage{import}
\usepackage{multirow}
\usepackage{booktabs}
\usepackage{pifont}
\usepackage{ulem}
\usepackage{algorithm}
\usepackage{algpseudocode}
\usepackage{tikz}
\usepackage{float}
\usetikzlibrary{arrows.meta, positioning, shapes.multipart, fit}

\usepackage{hyperref}

\renewcommand{\emph}[1]{\textit{#1}} 

\author[1]{Qian Wang}
\author[1]{Zahra Yousefijamarani}
\author[1]{Morgan Lindsay Heisler}
\author[1]{Rongzhi Gu}
\author[2]{Bai Xiaolong}
\author[2]{Shan Yizhou}
\author[1]{Wei Zhang}
\author[1]{Wang Lan}
\author[1]{Ying Xiong}
\author[1,$\dagger$]{Yong Zhang}
\author[1,$\dagger$]{Zhenan Fan}

\affil[1]{Huawei Technologies Canada Co., Ltd.}
\affil[2]{Huawei Technologies Co., Ltd.}
\renewcommand\Affilfont{\small}

\makeatletter
\renewcommand{\maketitle}{%
    \bgroup
    \setlength{\parindent}{0pt}
    \begin{adjustwidth}{0pt}{0pt}  
        \begin{flushleft}
            {\raggedright \titlefont \@title\par}%
            \vskip2pt
            {\raggedright \@author\par}%
        \end{flushleft}
    \end{adjustwidth}
    \vskip2pt

    {\abscontent}

    \begingroup
    \renewcommand{\thefootnote}{\fnsymbol{footnote}} 
    \footnotetext[2]{Corresponding authors: Yong Zhang <yong.zhang3@huawei.com>, Zhenan Fan <zhenan.fan1@huawei.com>.}
    \endgroup

    \thispagestyle{firststyle} 
    \egroup
}
\makeatother

\newcommand{\paper}{MEPIC}
\title{\vspace{-0.2in}\centering \paper{}: Memory Efficient Position Independent Caching\\for LLM Serving}

\begin{abstract}

Modern LLM applications such as deep-research assistants, coding agents, and Retrieval-Augmented Generation (RAG) systems, repeatedly process long prompt histories containing shared document or code chunks, creating significant pressure on the Key–Value (KV) cache, which must operate within limited memory while sustaining high throughput and low latency. Prefix caching partially alleviates some of these costs by reusing KV cache for previously processed tokens, but limited by strict prefix matching. Position-independent caching (PIC) enables chunk-level reuse at arbitrary positions, but requires selective recomputation and positional-encoding (PE) adjustments. However, because these operations vary across queries, KV for the same chunk diverges across requests. Moreover, without page alignment, chunk KV layouts diverge in memory, preventing page sharing. These issues result in only modest HBM savings even when many requests reuse the same content.

We present \paper{}, a memory-efficient PIC system that enables chunk KV reuse across positions, requests, and batches. \paper{} aligns chunk KV to paged storage, shifts recomputation from token- to block-level so only the first block is request-specific, removes positional encodings via Rotary Position Embedding (RoPE) fusion in the attention kernel, and makes remaining blocks fully shareable. These techniques eliminate most duplicate chunk KV in HBM, reducing usage by up to 2× over state-of-the-art PIC at comparable latency and accuracy, and up to 5× for long prompts, without any model changes.

\end{abstract}


\begin{document}
\begin{CJK*}{UTF8}{gbsn}

\maketitle

\newpage
\tableofcontents
\newpage

\section{Introduction}

Deep-research systems, coding agents, and Retrieval-Augmented Generation (RAG) pipelines all exhibit a common retrieval pattern: a disproportionate concentration of accesses on a small subset of documents, code files, or passages across many queries and many users. In RAG, the corpus may contain thousands of documents, yet a narrow head of highly relevant items dominates retrieval frequency ~\cite{zhao2024retrieval}. Coding agents demonstrate a similar skew, where a limited number of repositories, libraries, and header files account for the majority of lookups. Deep-research workflows follow the same trend—although they can issue broad web-scale searches, queries on related topics across users consistently converge on a small set of authoritative webpages and reports, reflecting the well-documented Zipfian retrieval behavior observed in large-scale search systems ~\cite{adamic2001web}.

This repeated‑context pattern is visible in production measurements. A recent study of Key–Value (KV) cache workloads at a large cloud provider ~\cite{wang2025kvcache} reports that: (i) KV reuse is highly skewed across requests, with a small fraction of prompts and prefixes accounting for most cache hits; (ii) reuses between independent, single‑turn requests are as important as reuses within multi‑turn chat sessions, indicating that many different users repeatedly hit the same context; and (iii) the cache size needed to achieve a near‑ideal hit ratio is moderate, consistent with the existence of a small “hot” set of prefixes and chunks. Together, these observations suggest that in real systems, the same document chunks are not only reused over time, but often reused by many active requests in overlapping time windows.

Modern multi‑turn, tool‑using agents ~\cite{xu2025llm} amplify this effect. Instead of issuing a single stateless completion, each user session spawns a sequence of calls as the agent plans, retrieves, and refines answers. These calls often reuse the same chunks (e.g., the same section of a report or the same code file) across steps. When many users interact with the system concurrently, the probability that multiple requests in the same batch refer to the same chunks in overlapping time windows becomes high. Ideally, the serving stack should treat these repeated chunks like a cache line in a CPU: compute their KV once, keep a single copy in High‑Bandwidth Memory (HBM), and let many requests alias it.

Without such cross‑request HBM reuse, each request that references a popular chunk must either (i) recompute its KV from scratch in HBM or (ii) reload a full copy of its KV from CPU/disk into HBM, even if tens or hundreds of other active requests reference the same content. Under high concurrency this multiplies HBM footprint roughly by the number of overlapping requests and forces the system to evict KV for other chunks or prefixes earlier than necessary. The result is lower effective capacity, more cache thrashing, and extra recomputation or data movement—all of which translate into higher tail latency and lower throughput.

Existing mechanisms only partially address this gap. Prefix caching in systems like vLLM ~\cite{2023vllm} reuses KV when multiple requests share an identical prefix, but its strict prefix‑matching requirement breaks as soon as the same chunks appear in different orders or at different positions within the prompt. Position‑independent caching (PIC) schemes~\cite{yao2025cacheblend,hu2024epic,yang2025kvlink} remove this restriction by allowing doc‑chunk KV to be reused at arbitrary positions in the prompt, at the price of recomputing some tokens and adjusting Positional Encodings (PE) to preserve accuracy. However, because recomputation and positional adjustment are performed independently per request, the resulting KV representations diverge across requests and cannot be page-aligned or shared in the paged KV cache. As a result, these approaches primarily reduce recomputation and off-GPU traffic, but do not fundamentally address HBM duplication when many requests reference the same chunks concurrently.



In this work, we present \paper{}, a memory‑efficient PIC system that lifts chunk reuse into the paged KV cache and makes chunk KV shareable in HBM both within and across batches. MEPIC (i) identifies reusable prompt regions and assigns them a canonical KV layout, padding each region to the system’s KV page size so identical regions map to identical KV blocks, (ii) moves selective recomputation from token‑level to block‑level by recomputing only the first block of each chunk per request while fusing RoPE into the attention kernel so the remaining blocks become position‑agnostic, and (iii) introduces a chunk cache coordinator that manages these canonical “clean” blocks in a shared HBM pool alongside vLLM’s existing prefix cache, resolving memory pressure via lazy LRU donation and integrating with LMCache for off‑device persistence. This design enables intra‑ and inter‑batch chunk KV sharing, significantly reducing HBM footprint and associated recomputation without requiring model changes.

Our contributions are as follows:
\begin{itemize}
    \item \textbf{Chunk-aware HBM KV management.} 
    We introduce a chunk cache coordinator that manages canonical chunk pages in a shared HBM pool alongside vLLM’s prefix cache, enabling coordinated allocation, reuse, and eviction under memory pressure. We perform deterministic, page-aligned chunk materialization to ensure that identical logical chunks map to identical HBM pages, and we employ lazy LRU-based eviction integrated with LMCache to support a remote persistence layer (CPU/Disk) for non-resident chunks.

    \item \textbf{Position-independent KV caching via fused RoPE attention.}
    We adopt a positional-encoding-free (NoPE) KV format in which attention states are stored without pre-applied rotary encodings. Positional information is instead injected on-the-fly within a fused RoPE attention kernel at execution time. This design decouples cached KV from absolute token positions, enabling deterministic chunk reuse regardless of
    where a chunk appears within the prompt.

    \item \textbf{System integration.}
    We integrate MEPIC into the vLLM + LMCache serving stack, demonstrating  that chunk‑level HBM reuse can be plugged into a production system with minimal engine changes while leveraging LMCache’s persistence layer as a remote chunk store.

    \item \textbf{Empirical benefits.}
    Through experiments, we show that \paper{} reduces HBM usage by up to 2× on multi-step RAG workloads and by more than 5× for long prompts, while achieving the same—or even better—accuracy and end-to-end latency than CacheBlend~\cite{yao2025cacheblend} and EPIC~\cite{hu2024epic}, \textit{without any model changes}.

\end{itemize}






Together, these contributions constitute the first design that enables page-aligned, position-independent chunk KV sharing directly in the paged HBM cache of a production LLM serving system.

\section{Background and Motivation}

\subsection{Background}

\subsubsection{LLM Serving and KV caching}

LLM serving systems execute transformer models on AI accelerators such as GPUs and NPUs, whose on-device HBM is scarce and costly. Inference is typically divided into two phases, \textbf{prefill} and \textbf{decode}. During the prefill phase, the model processes the full input prompt once and produces per-layer KV vectors for all prefix tokens. This phase often dominates the Time To First Token (TTFT), particularly in long-context workloads, because its computation grows linearly with prompt length. Once the prefix KV vectors are computed, the decode phase generates tokens auto-regressively. Each new token attends to the KV vectors stored from the prefix and from previously generated tokens. This \textbf{KV caching} pattern avoids redundant computation, but the resulting KV cache quickly becomes a major consumer of HBM and directly constrains batching capacity and throughput. To address this challenge and increase KV cache reuse, various KV reuse mechanisms have been proposed. A particularly effective approach is prefix caching.

\subsubsection{Prefix Caching}
The Prefix caching method~\cite{sglang2024, 2023vllm, gim2024prompt, gao2024cost} reuses the exact per-request prefix KV when multiple requests share identical leading tokens such as system prompts, few-shot exemplars, or fixed templates. Prefix reuse is simple and effective in many interactive services because it reduces both KV memory and TTFT for repeated beginnings. However, prefix caching depends on exact token-alignment and thus misses reuse opportunities when shared content appears in the middle of prompts or when chunks are reordered, as is common in retrieval-augmented generation and multi-document QA workflows. To enable cross-request reuse in these settings, more general position-independent reuse mechanisms have recently been proposed. 





\subsubsection{Position‑Independent Caching and Fusion Methods}
\label{pic}

\begin{figure*}
  \includegraphics[width=\textwidth]{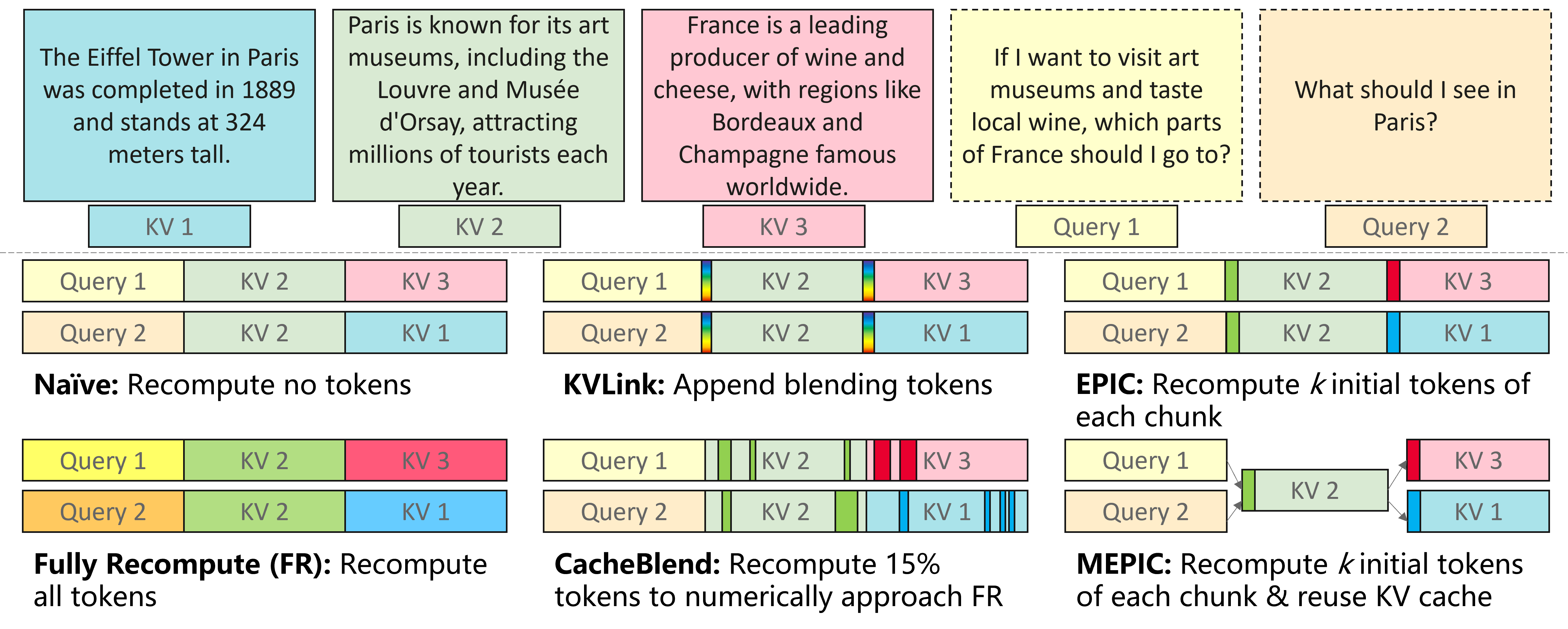}
  \caption{Comparison of PIC Algorithms. The area above the dashed line corresponds to the compile step, while the area below corresponds to the link step. The naive algorithm doesn't recompute any tokens, whereas the Fully Recompute Algorithm recomputes all tokens (highlights in darker colours). The four other PIC algorithms include KVLink, CacheBlend, EPIC, and MEPIC (our method). MEPIC enables cross‑request HBM reuse to reduce HBM usage thus improving system throughput. }
  \label{fig:baseline-comparison}
\end{figure*}

\paragraph{Algorithmic Taxonomy.} \emph{PIC}~\cite{hu2024epic} generalizes prefix reuse by precomputing KV representations for immutable text chunks so that a chunk can be reused regardless of its placement in the prompt. Figure~\ref{fig:baseline-comparison} illustrates how PIC methods differ from other approaches. Rather than recomputing all tokens (as in the Fully Recompute method) or none (as in the Naive method), PIC methods typically recompute only a subset of tokens. The key distinction lies in how each method determines which tokens to recompute. CacheBlend~\cite{yao2025cacheblend} is an early instantiation of this idea: it precomputes KV for retrieved chunks and selects the top $p$\% (usually 15\%) of chunk tokens order by \emph{K/V vector deviations}, recomputing their kv caches and applies RoPE realignment to the entire chunk. Subsequent methods introduce \emph{attention-aware} selection. A³~\cite{zhou20253} and KVShare~\cite{yang2025kvshare} derive top-$k$ tokens from attention scores, recomputing the regions most important to the query, while CacheClip~\cite{yang2025cacheclip} augments this with a lightweight predictor that classifies tokens into reuse or recompute classes. All of these methods perform \emph{dynamic} token selection and therefore incur query-dependent recomputation overhead. EPIC~\cite{hu2024epic}, in contrast, restores accuracy using a \emph{static sparsity} policy: for each reused chunk it deterministically recomputes the first $N$ tokens (e.g., 16 or 32), achieving accuracy comparable to dynamic methods while significantly reducing recomputation cost. KVLink~\cite{yang2025kvlink} occupies a different point in the design space: instead of relying on partial recomputation, it introduces small trainable link tokens to connect independently encoded segments, improving cross-chunk coherence but requiring model finetuning.

\paragraph{Operational overview of PIC systems.} Across these systems, the operational workflow of PIC-style KV reuse largely follows a common four-stage pipeline. (1) Each document chunk is precomputed offline and its KV cache is materialized and stored in HBM, CPU memory or on disk. (2) At query time, the system identifies reusable chunks and stages their cached KV into HBM-resident buffers. (3) A recomputation phase corrects positional mismatch by applying RoPE adjustments to all tokens in the reused chunk while selectively recomputing a subset of tokens determined by the method’s token-selection policy. (4) The resulting KV state, consisting of both reused and
recomputed tokens, is then written into the runtime paged-attention KV blocks
for subsequent execution. Where these systems diverge most significantly is in step (3).
Although all PIC methods must correct positional mismatch, they differ in how
recomputation is triggered and bounded, which directly affects per-request
compute cost and sensitivity to query-specific variation.

\paragraph{System-Level Gaps.} Despite this algorithmic diversity, existing works invest relatively little in system-level design, particularly in memory tiering, scheduling, and interactions between prefix caching and chunk caching. EPIC includes only a proof-of-concept implementation of a multi-tier KV store (HBM/CPU/Disk) and does not articulate how prefix KV and chunk KV should be jointly managed or balanced in realistic long-context workloads. CacheBlend’s paper does not discuss GPU memory management either; in LMCache’s open-source implementation, chunk KV resides exclusively in CPU/disk without an integrated HBM management strategy. Consequently, while current methods advance recomputation algorithms, their system-level implications such as KV residency, staging bandwidth, eviction, and alignment with paged attention remain underexplored. Critically, this gap is most severe because the effectiveness
of PIC techniques fundamentally depends on reliable \emph{in-HBM} chunk reuse. Without mechanisms that preserve chunk residency and ensure page-aligned, shareable KV layouts, reusable chunks are repeatedly evicted, misaligned, or duplicated across paged-attention blocks. This leads to low reuse hit rates, inflated memory consumption, unstable TTFT, and frequent CPU/disk restaging, effectively erasing the theoretical gains promised by selective recomputation. In practice, PIC algorithms only provide end-to-end benefits when chunk KV can persist reliably in HBM and be shared across concurrent requests. These observations motivate the need for canonical in-HBM chunk representations and dedicated HBM-resident management policies.


\subsection{Motivation}
\label{sec:motivation}
\subsubsection{Challenges}
PIC algorithms operate at the logical-KV level, but HBM-resident reuse requires new serving-system mechanisms. In real serving stacks such as vLLM, enabling high-throughput, multi-tenant chunk reuse requires a canonical, page-aligned, position-independent representation of chunk KV, along with scheduling and memory-management support. Achieving these goals, however, faces three system-level challenges.

\paragraph{Challenge \ding{182}: No native support for in-HBM chunk management.}
Existing PIC systems operate entirely outside the inference engine: chunks reside in CPU or disk tiers, and their KV is staged into HBM only on demand. In contrast, serving systems such as vLLM pre-allocate almost all HBM as a single paged-attention KV store. The attention kernels directly read from this internal KV store, and do not accept pointers to external memory regions; nor does vLLM expose any mechanism to reserve, index, allocate, or evict chunk-level KV blocks.

As a result, maintaining chunk KV outside of vLLM is not viable:  
(1) external chunks cannot be consumed by the attention kernels without intrusive modifications or forking them;  
(2) copying external chunk KV into vLLM's paged-KV store defeats reuse by adding transfer overhead; and  
(3) without residency control inside the paged-KV pool, chunk KV would be overwritten by prefix KV or normal request traffic.

Therefore, enabling in-HBM chunk reuse requires a fundamental architectural extension to the serving engine itself:
\begin{itemize}
    \item chunk-aware indexing and scheduling,  
    \item chunk-level block allocation, reclamation, and offloading,  
    \item unified management of prefix KV and chunk KV within a shared block pool.  
\end{itemize}

Without such a subsystem, chunk KV cannot remain resident in HBM, cannot be shared across requests, and cannot be accessed by paged-attention execution in a performant or maintainable way.

\paragraph{Challenge \ding{183}: Lack of canonical, page-aligned block placement for chunks.}
Even if chunk KV can reside in HBM, it cannot be shared across requests unless its physical layout in the paged-attention KV cache is \emph{canonical}—i.e., it must occupy the same sequence of KV blocks in every request. Two factors prevent this.

\textbf{(1) Block misalignment.}
The starting offset of a chunk depends on the total prompt length preceding it, which varies across requests. Because vLLM assigns KV blocks in fixed-size pages, a chunk that begins at different offsets will map to different physical blocks. Even if two requests reference the exact same chunk, their KV will occupy different page boundaries, making block-level reuse impossible.  
For example, with a block size of 16 tokens, a chunk starting at offset 32 in one request and offset 35 in another will occupy entirely different KV blocks despite being semantically identical.

\textbf{(2) Inconsistent recomputation patterns.}
PIC selectively recomputes a subset of chunk tokens to restore precision. If these recomputed tokens are distributed across different positions in different requests, they introduce request-specific ``dirty'' blocks. Because reuse in paged-attention occurs at block granularity, a single recomputed token in a block renders the entire block non-shareable.  
With random or dynamically sparse selection, it is possible for every block of a chunk to become dirty in one request, eliminating the possibility of cross-request reuse.

Therefore, supporting chunk reuse requires enforcing block-aligned chunk placement and establishing predictable clean/dirty boundaries (e.g., statically recomputing only the first block). Without canonical block layouts, reused chunks cannot share KV blocks even if their token sequences are identical.

\paragraph{Challenge \ding{184}: Positional encoding breaks canonical chunk reuse.}
Even with canonical, block-aligned chunk KV, reuse is still impossible if positional encoding has already been applied. In RoPE-based models, the key and value representations are functions of absolute token positions. Consequently, the same chunk appearing at different offsets in different requests produces distinct K/V tensors, even when the token sequence and block layout are identical.

Because positional encoding is embedded directly into the stored KV, cached chunk KV becomes inherently request-specific and cannot serve as a canonical, reusable representation. Any attempt to reuse such KV would require recomputing or adjusting positional encoding, reintroducing computation overhead and breaking reuse guarantees.

As a consequence, positional dependence emerges as a fundamental obstacle to
cross-request chunk reuse: without addressing how positional information
interacts with cached KV, chunk reuse cannot be made both correct and efficient
in long-context, multi-tenant serving environments.

\paragraph{Why these challenges matter.}
Taken together, these challenges expose a fundamental gap between PIC algorithms and practical LLM serving systems. While PIC identifies which tokens \emph{could} be reused, it provides no system support for \emph{how} chunk KV can be represented, placed, and preserved inside HBM. Without (1) native in-HBM chunk management, (2) canonical, page-aligned KV placement, and (3) position-independent KV representations, chunk KV cannot remain resident, cannot be shared at block granularity, and cannot survive eviction pressure in multi-tenant environments. Therefore, reuse degenerates into thrashing, KV duplication, unpredictable HBM utilization, and unstable hit rates.

These limitations make chunk reuse not an algorithmic problem alone, but a system design problem. This motivates a dedicated in-HBM chunk-reuse subsystem, tightly integrated with vLLM’s paged-attention architecture, that treats chunk KV placement, alignment, and residency as first-class concerns.

\section{Design}
\label{sec:design}

MEPIC addresses these challenges by bridging the gap between PIC algorithms and practical LLM serving systems, making chunk-level KV reuse practical. Rather than introducing new recomputation algorithms,
MEPIC focuses on how reusable chunk KV can be represented, placed, and preserved
within the serving system so that existing PIC techniques can deliver their
intended benefits under realistic, multi-tenant workloads.

At a high level, MEPIC treats chunk KV as a first-class, HBM-resident object and
integrates its management directly into vLLM’s paged KV abstraction. The system
is organized around two execution paths already present in modern serving
stacks: a \emph{scheduling path}, which determines KV placement and residency
using metadata only, and a \emph{computation path}, which materializes KV and
executes attention. MEPIC extends both paths to support chunk-aware reuse while
preserving vLLM’s existing execution interface and attention semantics.

This section first presents a system overview (\S\ref{sec:overview}), then
describes the scheduling path responsible for chunk identification, placement,
and eviction (\S\ref{sec:scheduling}), followed by the computation path that
performs selective recomputation and fused-RoPE attention execution
(\S\ref{sec:compute}).

\subsection{System Overview}
\label{sec:overview}


MEPIC integrates chunk-level KV reuse into a vLLM+LMCache serving stack by
extending the standard scheduling and execution workflow, without introducing
a separate KV pool or departing from vLLM’s paged-attention execution model.

\begin{figure}[t]
    \centering
    \includegraphics[width=0.7\columnwidth]{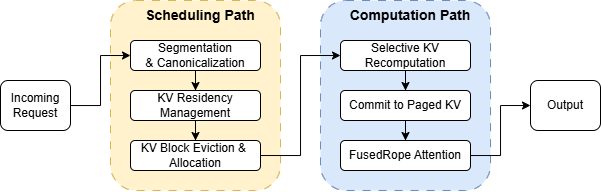}
    \caption{MEPIC system overview integrated into a vLLM/LMCache serving stack.
The scheduling path constructs a chunk-aware KV placement plan
within vLLM’s paged KV store, and the computation path follows this plan to
recompute necessary tokens and execute attention with fused RoPE.}
    \label{fig:mepic_system}
\end{figure}

As illustrated in Figure~\ref{fig:mepic_system}, incoming requests follow the
same two-path structure as vanilla vLLM. In the \emph{scheduling path}, MEPIC
augments request processing with chunk-aware segmentation and cache coordination,
constructing a deterministic placement plan within vLLM’s paged KV store. The
output interface remains unchanged: a (padded) token sequence and a mapping from
tokens to paged KV blocks, now enriched to reflect chunk sharing and cache
provenance.

The \emph{computation path} consumes this output without additional metadata.
Chunk and prompt regions are inferred implicitly from the padding pattern, and
execution proceeds using vLLM’s paged-attention backend. Selective recomputation
and Fused-RoPE attention are applied during execution, enabling
reusable chunk KV to be shared safely across requests.

\subsection{Scheduling Path: Chunk-Aware KV Management}
\label{sec:scheduling}

\begin{figure}[t]
\centering
\includegraphics[width=0.7\columnwidth]{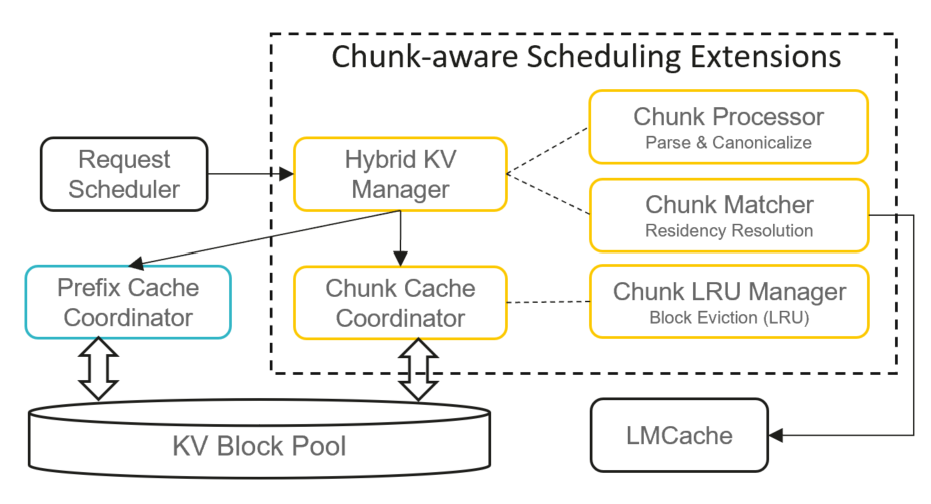}
\caption{Scheduling components introduced by MEPIC for chunk-aware KV management.
The Hybrid KV Manager coordinates prefix and chunk handling across shared HBM KV
blocks, while specialized components enforce canonical chunk alignment, resolve
cache residency, and manage allocation and eviction across local and remote
tiers. Together, these components integrate chunk KV as a first-class object
into vLLM’s scheduling path without changing its execution interface.}

\label{fig:sched-components}
\end{figure}

The scheduling path constructs a chunk-aware KV management substrate inside the serving engine. Operating entirely on metadata and cache state, it determines how a request’s KV state is assembled in vLLM’s paged KV store before any token-level computation. Figure~\ref{fig:sched-components} illustrates \paper{}'s additional scheduling components and their interactions, which preserve vLLM’s standard scheduling interface.

\subsubsection{Segmentation and Canonicalization}
\label{sec:segmentation}

The scheduling path first establishes a canonical
representation for reusable KV segments through the \emph{Chunk Processor}. Without deterministic segmentation and block-aligned placement, identical chunks at different offsets would map to different paged-KV layouts and could not be safely shared.

\paper{} partitions each request into \emph{chunk segments} (immutable, reusable content) and \emph{prompt segments} (request-specific content). Lightweight segment markers delineate these regions, and segmentation occurs entirely in the scheduler without token-level computation.

To enforce deterministic placement, segments are padded at block granularity: chunk segments receive \emph{leading} padding, prompt
segments receive \emph{trailing} padding, and the final prompt segment is left unpadded. 
This asymmetric scheme ensures each reusable chunk begins at a block boundary. The padding pattern also implicitly encodes segment type, allowing the computation path to distinguish chunks from prompts without additional metadata.


\begin{figure}[t]
    \centering
    \includegraphics[width=0.7\columnwidth]{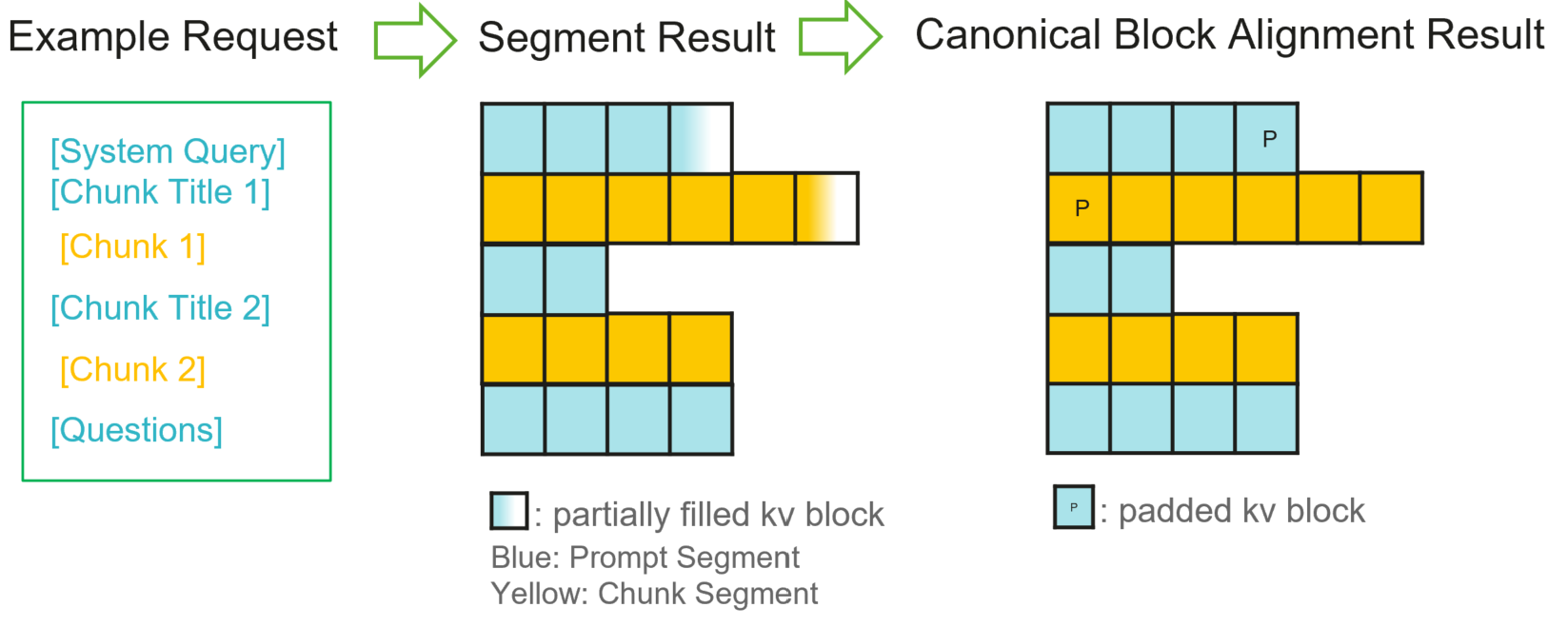}
    \caption{
    Segmentation and canonical block alignment in MEPIC.
Padding enforces a canonical, block-aligned KV layout, allowing identical chunk
segments to reuse the same KV blocks across requests.
    }
    \label{fig:Segmentation}
\end{figure}

Figure~\ref{fig:Segmentation} shows a concrete example of this process,
highlighting how asymmetric padding yields canonical, block-aligned KV layouts
for reusable chunks. In the figure, chunk segments are shown in yellow, while prompt segments are shown in blue.
The first two segments partially fill their KV blocks and therefore require padding for alignment.
Therefore, the chunk segment is padded at the beginning, whereas the prompt segment is padded at the end.

\subsubsection{Chunk-Aware KV Residency Management}
\label{sec:residency}

After segmentation, the scheduling path resolves the
HBM residency status of each segment through the \emph{Chunk Matcher}. This information is then given to the \emph{Hybrid KV Manager} to coordinate with the Prefix and Chunk Cache Coordinators to classify segments as HBM-resident or non-resident and to identify remote copies when available. For each segment, the Hybrid KV Manager consults the Prefix Cache Coordinator and the Chunk Cache Coordinator to identify existing KV blocks in HBM, as well as remote chunk entries maintained by LMCache. These lookups classify
segments as HBM-resident or non-resident and record the availability of remote copies for non-resident segments.



Explicit residency resolution ensures that subsequent allocation, eviction, and recomputation policies are applied consistently, supporting efficient reuse under memory pressure. 


\subsubsection{Eviction and Allocation under Pressure}
\label{sec:lru}

Once residency is known, the scheduler ensures sufficient HBM KV blocks. Free blocks are allocated directly; if memory is constrained, the \emph{Chunk LRU Manager} reclaims zero-reference chunk KV blocks. Prefix KV blocks are never evicted to maintain correctness and predictable latency.


\paragraph{KV Block Allocation.}

\paper{} manages prefix and chunk KV within a shared paged-KV block pool. The padded token sequence of each chunk is hashed to determine its identity, in contrast to vLLM’s prefix cache, which hashes KV blocks independently. 

Upon allocation, the \emph{Chunk Cache Coordinator} acquires the required KV blocks from the shared pool and registers the chunk with the \emph{Chunk LRU Manager}, initializing its reference count to one. Selective recomputation (section~\ref{sec:kvrecompute}) further influences allocation semantics: the first KV block is recomputed via the prefix cache, ensuring boundary correctness, while remaining blocks form the canonical, shareable chunk KV managed by the chunk cache. Allocation updates reference counts and chunk LRU state.



\begin{figure}[t]
    \centering
    \includegraphics[width=0.7\columnwidth]{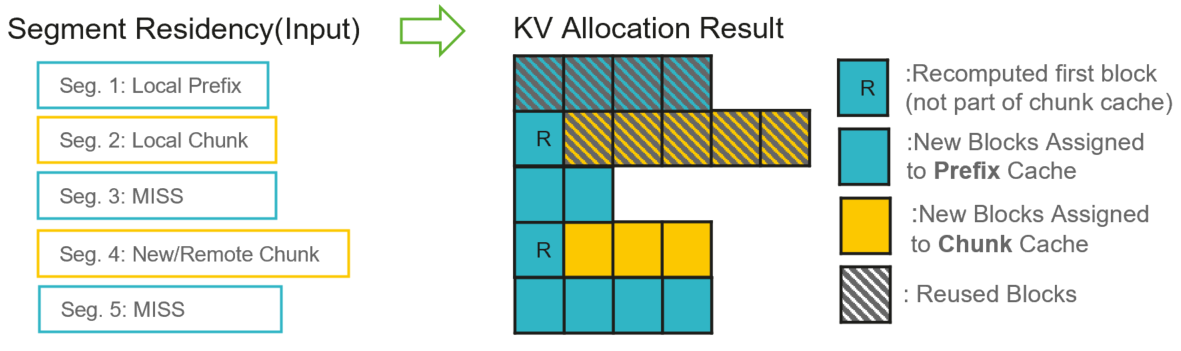}
    \caption{An example of KV block allocation following segment residency
classification. Based on per-segment residency, reusable KV blocks are shared,
while non-resident segments are assigned newly allocated blocks from the shared
HBM pool. For chunk segments, the first KV block is deterministically
recomputed and allocated via the prefix cache, while the remaining blocks form
canonical, shareable chunk KV managed by the chunk cache.}
    \label{fig:allocation-example}
\end{figure}

Figure~\ref{fig:allocation-example} illustrates how per-segment residency
classification is translated into KV block allocation decisions.

\paragraph{Eviction Policy.}

Eviction decisions are guided by the \emph{Chunk LRU Manager}, which tracks usage of HBM-resident chunk KV objects and selects eviction candidates under memory pressure. Reference-counted chunks with zero active references are eligible for eviction, while prefix KV blocks are never evicted to maintain correctness and predictable latency. Object-level eviction preserves canonical block alignment and ensures predictable reuse behavior under memory pressure.

Algorithm~\ref{alg:mepic-sched} summarizes the scheduling path control logic,
highlighting the ordering of these decisions and the admission check performed
before KV block allocation.

\begin{algorithm}[t]
\caption{MEPIC Scheduling Path}
\label{alg:mepic-sched}
\begin{algorithmic}[1]

\State \textbf{Input:} Request token sequence $T$ with segment markers
\State \textbf{Output:} Padded tokens $T_{\text{align}}$ and paged KV block mapping, or \textsc{Reject}

\Statex
\State \textbf{// Segmentation and Canonicalization}
\State Partition $T$ into segments $\{S_1,\ldots,S_m\}$ based on segment markers
\For{each segment $S_i$}
    \State Determine segment type (chunk or prompt)
    \State Apply padding to enforce block-aligned canonical layout
\EndFor
\State Construct padded token sequence $T_{\text{align}}$

\Statex
\State \textbf{// Residency Resolution}
\For{each segment $S_i$}
    \State Resolve HBM residency and remote availability
\EndFor

\Statex
\State \textbf{// Admission Check}
\State Estimate number of KV blocks required by non-resident segments
\If{insufficient free blocks and evictable chunk KV cannot satisfy demand}
    \State \Return \textsc{Reject}
\EndIf

\Statex
\State \textbf{// KV Block Allocation}
\For{each segment $S_i$}
    \State Assign paged KV blocks according to residency and segment type
\EndFor

\State \Return $T_{\text{align}}$ and KV block mapping

\end{algorithmic}
\end{algorithm}

\subsection{Computation Path}
\label{sec:compute}

The computation path materializes the required KV state and executes attention computation using the scheduling output. It remains agnostic to residency and allocation decisions.


\subsubsection{Selective KV Recomputation}
\label{sec:kvrecompute}

As mentioned in Section~\ref{sec:segmentation}, segment types are inferred from the padding pattern. Prompt segments are fully recomputed, as their content is request-specific and cannot be safely shared across requests. For chunk segments, newly encountered chunks are fully recomputed, while cached chunks require recomputation of only the first KV block, with the remaining blocks reused as canonical KV. This captures context-dependent attention at chunk boundaries, minimizes compute overhead, and preserves the accuracy benefits of chunk-level reuse.




\subsubsection{Commit to Paged KV}
\label{sec:kvcommit}

Recomputed KV vectors are then written into the assigned paged KV blocks. As the scheduling path has already determined a deterministic mapping from tokens to KV blocks, this step performs a direct write into the assigned pages without additional indirection or metadata translation.

 Block-aligned segmentation and padding ensures that both recomputed and reused blocks occupy stable locations, enabling safe sharing without additional metadata translation. This commit step completes KV materialization and prepares the paged KV store for subsequent attention execution.


\subsubsection{Fused RoPE Attention}
\label{sec:rope}


To enable position-independent reuse of chunk KV, \paper{} separates RoPE from KV cache storage.
In standard vLLM execution, RoPE is applied to keys and values \textit{before} they are written to the paged KV cache. As a result, cached KV entries are tied to a specific absolute position in the prompt and cannot be reused at different offsets.

\paper{} instead stores KV in a positional-encoding-free (NoPE) format, omitting RoPE during KV materialization. Positional encoding is deferred until attention computation. When the attention kernel loads NoPE KV blocks from HBM into on-device buffers, it applies the appropriate rotary offsets on the fly within a fused RoPE–attention operator, immediately before computing attention scores.

This design allows the same canonical KV blocks to be reused at different prompt offsets across requests without recomputation. Because RoPE is fused into the attention operator and applied to on-device data, the approach avoids additional memory traffic and incurs negligible overhead compared to standard attention, while eliminating the need to maintain position-specific KV copies in HBM.

\section{Evaluation}
\subsection{Experiment Setup}

We implement our system on top of vLLM~\cite{2023vllm} and LMCache~\cite{cheng2025lmcache},
extending the existing KV management pipeline to support chunk-aware KV residency,
deterministic page-aligned chunk placement, selective KV recomputation, and
position-independent KV reuse.
The implementation incorporates deterministic, page-aligned chunk materialization,
selective recomputation, and NoPE KV caching. All
baselines are also implemented within the same framework and
evaluated under identical scheduling, batching, and execution settings to ensure
fair comparison.

\textbf{Datasets.} We evaluate \paper{} on four question-answering and reading comprehension datasets: SQuAD~\cite{rajpurkar2016squad}, NewsQA~\cite{trischler2017newsqa}, NarrativeQA~\cite{kovcisky2018narrativeqa}, and emrQA~\cite{pampari2018emrqa}. Paragraphs within these datasets are treated as semantic chunks, as they naturally correspond to reusable and semantically coherent document units, and we sample 300 requests from each dataset to construct evaluation workloads.

\textbf{Workload Characteristics.}
Across all evaluated datasets, a substantial fraction of input tokens correspond to reusable document chunks rather than request-specific content. On NewsQA, which represents the most challenging workload, an average of 61.4\% of tokens are reused, with the remaining tokens reflecting question-specific prefixes and formatting, or chunks that were not reused. SQuAD exhibits higher reuse at 84.9\% on average, while NarrativeQA and emrQA show heavy reuse, with mean reusable fractions of 93.9\% and 98.2\%, respectively. 

In absolute terms, requests contain between 1.4K and 2.2K tokens on average, while only 29–521 tokens per request require recomputation depending on the dataset. These results highlight both the diversity of reuse patterns across workloads and the presence of a significant reusable core even in less favorable cases such as NewsQA. This diversity allows our evaluation to stress-test chunk-aware KV reuse under varying degrees of reuse intensity rather than relying on uniformly cache-friendly inputs.

\begin{table}[t]
\centering
\small
\begin{tabular}{l|ccc}
\hline
Dataset & Mean tokens & Mean reuse (\%) & Mean recomputed \\
\hline
NewsQA      & 1518        & 61.4           & 521   \\
SQuAD       & 2224        & 84.9           & 321   \\
NarrativeQA & 1435        & 93.9           & 95     \\
emrQA       & 1632        & 98.2           & 29     \\
\hline
\end{tabular}
\caption{\textbf{Workload Characteristic Summary.} Mean request length, fraction of reused tokens, and tokens requiring recomputation per request across datasets. The workloads span a wide range of reuse intensity, from moderate reuse in NewsQA to near-complete reuse in NarrativeQA and emrQA.}
\label{tab::workload}
\end{table}







\textbf{Models.}
All experiments use the Mistral-7B-Instruct-v0.3~\cite{jiang2023mistral7b} model. Prior work on PIC techniques, including CacheBlend and EPIC,
has demonstrated that selective recomputation and position-independent KV reuse
generalize across a wide range of model sizes and architectures. Accordingly, we
focus on a single representative open-source model so that observed performance
differences primarily reflect the impact of the chunk-aware KV caching and
memory management mechanisms introduced by our system, rather than differences
in model architecture or scale.


\textbf{Hardware.} Experiments are conducted on Ascend 910B NPUs~\cite{liao2021ascend}, each equipped with 64~GB of HBM. The system leverages HBM-resident KV storage to measure the effectiveness of chunk reuse and page-aligned memory management.


\textbf{Baselines.} We compare our approach against two prior position-independent caching systems: EPIC~\cite{hu2024epic} and CacheBlend~\cite{yao2025cacheblend}. These baselines represent state-of-the-art methods for selective KV recomputation and position-independent caching. In our evaluation, we use CacheBlend with a 15\% recomputation ratio and EPIC configured to recompute 16 tokens, following the recommended settings reported in their respective papers.


\textbf{Metrics.} We evaluate system performance using three complementary metrics. First, \textit{latency} includes queueing delay, prefill, and decoding, and is measured end-to-end: from request submission to completion of the final generated token. 
Lower latency indicates more efficient computation, reflecting the effectiveness of KV caching and chunk reuse. Second, \textit{model accuracy} is assessed using dataset-specific metrics. For SQuAD, NewsQA, and emrQA, we report F1 score, which measures the overlap between predicted and ground-truth answers. For NarrativeQA, we use the Rouge-L score to evaluate the similarity of generated summaries to reference answers. Higher values indicate better performance. Third, \textit{HBM usage} is measured per accelerator and quantifies the memory footprint of KV caches in memory, excluding model weights, activations, and optimizer state. We report both peak usage over the request lifetime and average usage across all active requests, providing insight into the effectiveness of chunk-level cache sharing, page alignment, and NoPE KV reuse.

\begin{table*}[t]
\centering
\small
\resizebox{\textwidth}{!}{%
\begin{tabular}{l|ccc|ccc|ccc}
\hline
\textbf{Dataset} &
\multicolumn{3}{c|}{\textbf{Latency (s) ($\downarrow$)}} &
\multicolumn{3}{c|}{\textbf{Peak HBM Usage (\%) ($\downarrow$)}} &
\multicolumn{3}{c}{\textbf{Score ($\uparrow$)}} \\
&
\textbf{Ours} & \textbf{CacheBlend} & \textbf{EPIC} &
\textbf{Ours} & \textbf{CacheBlend} & \textbf{EPIC} &
\textbf{Ours} & \textbf{CacheBlend} & \textbf{EPIC}
\\ \hline
SQuAD &
$116.03 \pm 0.65$ & $119.41 \pm 0.88$ & $\mathbf{114.73 \pm 1.55}$ &
$\mathbf{27.67 \pm 0.05}$ & $54.47 \pm 0.25$ & $54.13 \pm 0.21$ &
$\mathbf{0.74 \pm 0.01}$ & $0.73 \pm 0.02$ & $0.72 \pm 0.02$ \\
NewsQA &
$\mathbf{112.36 \pm 1.10}$ & $117.00 \pm 5.07$ & $115.39 \pm 3.51$ &
$\mathbf{36.43 \pm 0.56}$ & $45.97 \pm 0.24$ & $45.43 \pm 0.09$ &
$0.65 \pm 0.02$ & $\mathbf{0.66 \pm 0.01}$ & $\mathbf{0.66 \pm 0.01}$ \\
NarrativeQA &
$\mathbf{97.71 \pm 0.78}$ & $104.72 \pm 4.11$ & $100.06 \pm 2.90$ &
$\mathbf{29.67 \pm 0.12}$ & $50.40 \pm 0.00$ & $50.40 \pm 0.17$ &
$0.60 \pm 0.01$ & $0.61 \pm 0.01$ & $\mathbf{0.61 \pm 0.00}$ \\
emrQA &
$\mathbf{105.67 \pm 1.23}$ & $110.97 \pm 1.33$ & $109.85 \pm 0.53$ &
$\mathbf{23.20 \pm 0.00}$ & $37.37 \pm 0.05$ & $37.40 \pm 0.08$ &
$\mathbf{0.39\pm 0.02}$ & $\mathbf{0.39\pm 0.02}$ & $\mathbf{0.39\pm 0.02}$ \\
\hline
\end{tabular}}
\caption{Comparison of total end-to-end latency, HBM usage, and score (F1 score: SQuAD, NewsQA, and emrQA; Rouge-L score: NarrativeQA) across four datasets for three baselines: Our system, CacheBlend, and EPIC.}
\label{tab::results}
\end{table*}

\begin{figure*}[t]
  \centering
  \includegraphics[width=\textwidth]{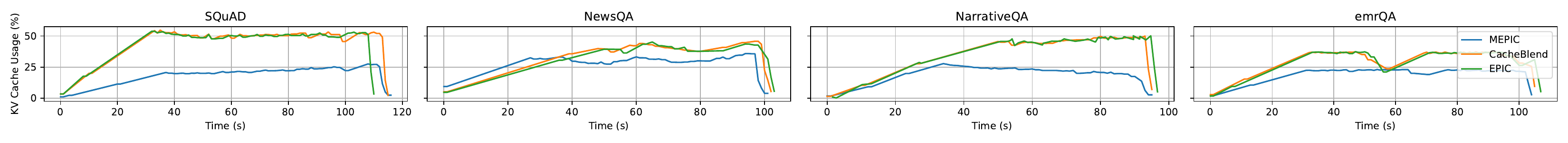} 
  \caption{HBM usage over time for each dataset (SQuAD, NewsQA, NarrativeQA, emrQA) comparing CacheBlend, EPIC, and \paper{}. The plots illustrate how chunk-aware KV reuse reduces memory consumption during inference.}
  \label{fig:usageplot}
\end{figure*}

\subsection{Baseline Comparison}



We first evaluate the effect of chunk-aware KV caching on model accuracy, HBM usage, and latency. Table~\ref{tab::results} compares these metrics for our system against EPIC and CacheBlend across four datasets: SQuAD, NewsQA, NarrativeQA, and emrQA. Despite introducing selective recomputation and NoPE KV caching, our system achieves comparable or slightly higher accuracy than the baselines. These results indicate that recomputing only the first KV block of cached chunks is sufficient to preserve model fidelity while enabling aggressive KV reuse.

In terms of latency, our method consistently demonstrates improvements across most datasets and workloads. This highlights the benefit of chunk-aware caching in accelerating end-to-end inference without altering the underlying model.

Regarding HBM usage, our system significantly reduces memory consumption by avoiding unnecessary KV storage. Across datasets, peak HBM usage is up to 2× lower than both CacheBlend and EPIC. This reduction not only allows larger prompts to fit in the available high-bandwidth memory but also reduces memory pressure, enabling more efficient handling of concurrent requests. While Table~\ref{tab::results} reports only the peak HBM usage, Figure~\ref{fig:usageplot} provides a more detailed view of HBM consumption over time. These plots illustrate how our chunk-aware KV reuse maintains lower memory usage throughout the inference process, and how memory utilization patterns vary across datasets with different amounts of reusable content.

Minor differences across datasets reflect the proportion of each prompt consisting of reusable, document-like content versus request-specific tokens, which in turn determines the fraction of KV that must be recomputed.







\subsection{HBM Usage and Latency Under Varying QPS}

\begin{figure}[t]
\centering
  \includegraphics[width=0.8\columnwidth]{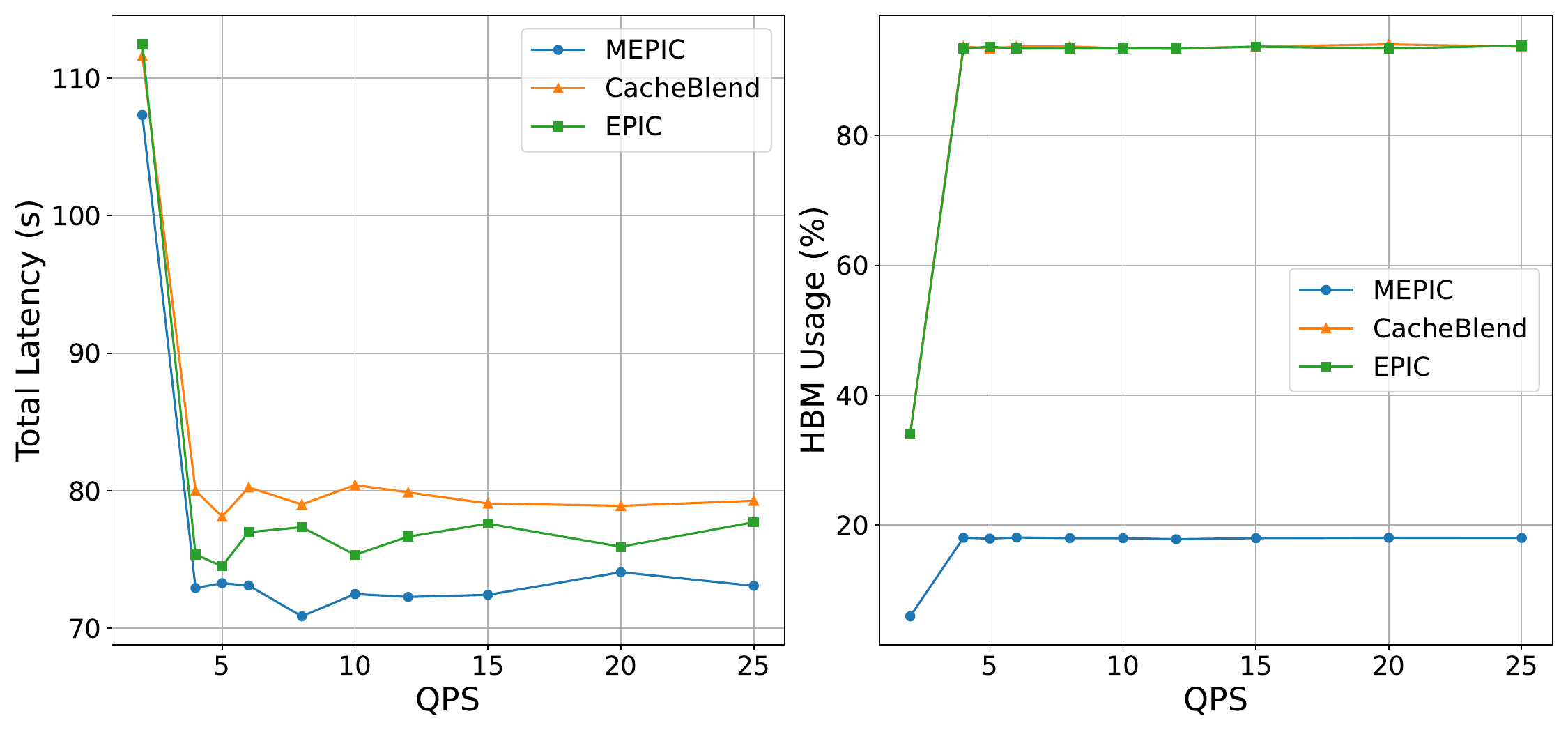} 
  \caption{HBM usage and end-to-end latency across QPS for CacheBlend, EPIC, and \paper{}. Our chunk-aware KV reuse reduces HBM consumption and improves latency across load levels.}
  \label{fig:qps}
\end{figure}

To evaluate system efficiency under varying load conditions, we measure HBM utilization and end-to-end latency across query rates ranging from 2 to 25 queries per second (QPS). We run 200 requests using Mistral-7B-Instruct-v0.3. As shown in Figure~\ref{fig:qps}, our chunk-aware KV management significantly reduces peak HBM usage compared to EPIC and CacheBlend by enabling broad reuse of canonical chunk KV blocks. It lowers HBM usage by 5.74× relative to CacheBlend and 5.25× relative to EPIC. Latency is also consistently lower across all QPS levels due to reduced prefill computation, achieving 9.1\% lower latency than EPIC and 11.48\% lower latency than CacheBlend.

EPIC exhibits lower latency than CacheBlend because it recomputes a fixed number of tokens per chunk, whereas CacheBlend recomputes approximately 15\% of tokens. Overall, these results demonstrate that our design improves memory efficiency while sustaining low TTFT and robust performance under production-scale load.




\subsection{HBM Usage and Latency Under Varying Context Lengths}

\begin{figure}[t]
\centering
  \includegraphics[width=0.8\columnwidth]{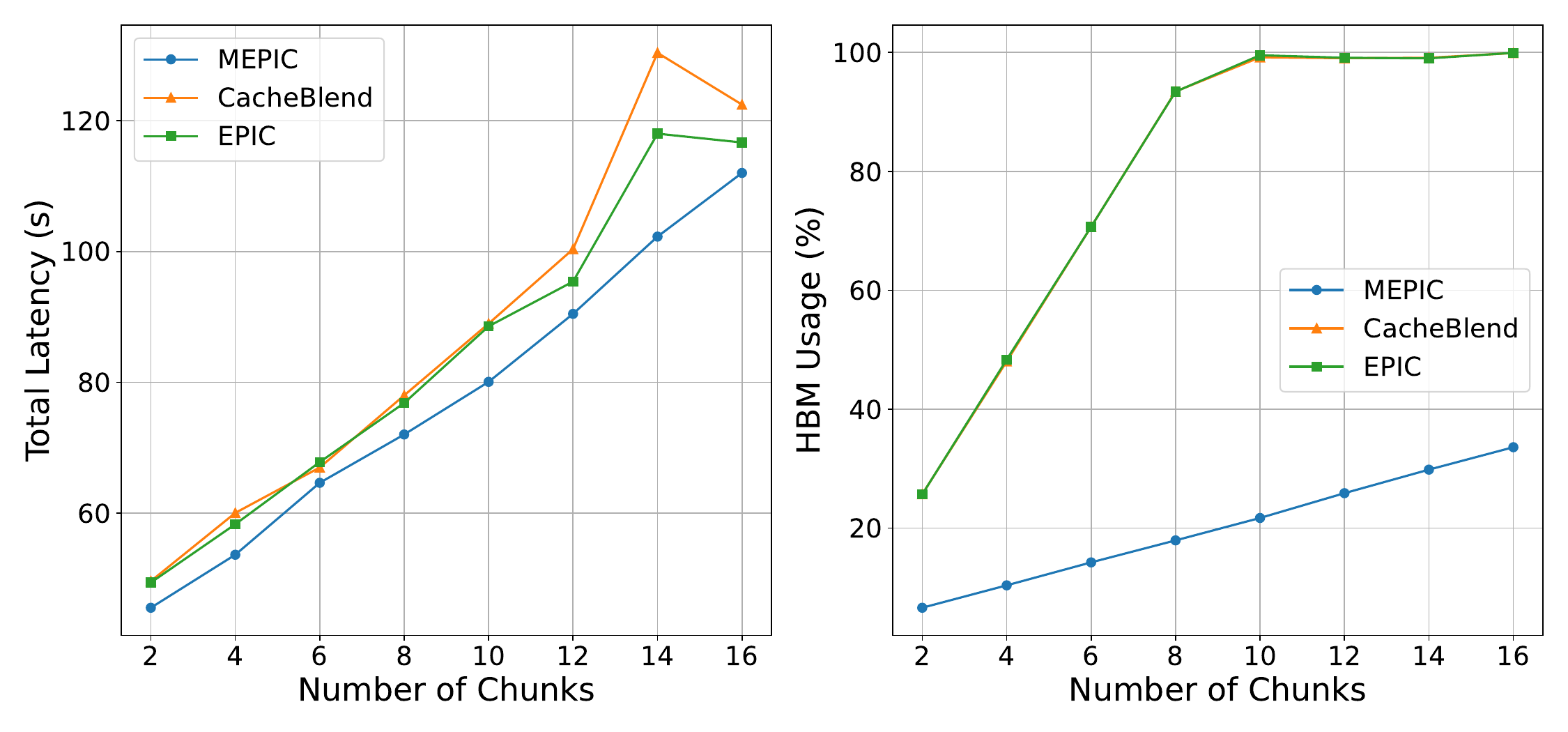}
  \caption{{HBM usage and end-to-end latency across increasing context lengths for CacheBlend, EPIC, and \paper{}. Our method consistently maintains lower HBM footprint and latency, even for very long prompts. }}
  \label{fig:long-context-ablation}
\end{figure}

We evaluate how increasing prompt length, expressed as the number of chunks, affects system performance. Using synthetically extended inputs, we measure HBM usage and end-to-end latency across three configurations: CacheBlend, EPIC, and \paper{}. We run 200 requests using Mistral-7B-Instruct-v0.3 at QPS=15, varying the number of chunks from 2 to 16. Figure~\ref{fig:long-context-ablation} shows that \paper{} consistently achieves substantially lower HBM consumption across all context sizes. As the context grows, EPIC and CacheBlend quickly saturate HBM capacity, preventing them from batching additional requests, whereas \paper{} remains below 40\% usage. Across the evaluated chunk sizes, \paper{} uses 2.97× to 5.21× less HBM than existing methods. Latency follows a similar trend: \paper{} maintains consistently lower end-to-end latency due to reduced prefill recomputation and minimized redundant KV materialization. While these inputs are synthetically extended, they allow controlled evaluation of KV growth behavior and reuse efficiency under increasing context lengths, isolating memory effects that would otherwise be confounded by dataset-specific semantics.





\subsection{Summary and Analysis}

Our experiments demonstrate that chunk-aware KV caching effectively balances memory efficiency, latency, and accuracy across a range of tasks and operating conditions. The baseline comparison shows that our system preserves or slightly improves model output quality relative to EPIC and CacheBlend, confirming that selective recomputation of the first block within each chunk, combined with on-the-fly positional encoding, maintains output fidelity. Minor variations in F1 and Rouge-L scores across datasets are consistent with differences in the proportion of mutable tokens per prompt, highlighting the relevance of chunk composition for caching strategies. Notably, while NarrativeQA and emrQA exhibit high reuse ratios, NewsQA presents a substantially lower reuse regime, yet our approach continues to deliver memory and latency improvements, indicating robustness beyond near-ideal caching scenarios.

When evaluating performance under varying query rates, our system substantially reduces peak HBM usage and consistently lowers latency. By reusing canonical chunk KV blocks, our method achieves a 5× reduction in memory footprint compared to the baselines while simultaneously reducing end-to-end latency by 9–11\%. This indicates that chunk-level caching not only alleviates memory pressure but also accelerates response times by minimizing redundant prefill computations. In contrast, EPIC’s fixed-token recomputation and CacheBlend’s partial recomputation lead to higher HBM usage and less predictable latency, particularly under increased load.

Performance under extended context lengths further underscores the scalability of our approach. Whereas EPIC and CacheBlend quickly saturate HBM as prompt length grows, our system maintains memory utilization below 40\%, enabling efficient batching of long-context requests. End-to-end latency similarly remains lower across all context sizes, demonstrating that our method effectively mitigates the computational overhead associated with large KV states. These results suggest that chunk-aware KV management is particularly advantageous in scenarios with long or highly variable prompts, where memory constraints are a limiting factor for throughput and latency.

Taken together, these findings highlight the practical utility of chunk-aware KV caching in production LLM inference. Although our experiments focus on a single representative model and accelerator, the evaluated mechanisms operate at the KV cache and attention-kernel level and are orthogonal to model architecture and parameter scale, suggesting applicability to other transformer-based LLMs.
Our approach achieves a favorable balance between accuracy, memory efficiency, and latency, outperforming existing baselines under diverse workloads and scaling conditions. Beyond the immediate performance gains, this strategy provides a foundation for further optimizations, such as dynamic chunk sizing or adaptive recomputation policies, which could further enhance efficiency in real-world deployments.

\section{Discussion}

Rather than proposing isolated optimizations, \paper{} makes a set of
interdependent design choices to enable practical chunk-level KV reuse in
production LLM serving systems. This section discusses the key trade-offs
underlying these choices and their implications for long-context, multi-tenant
inference.

\paragraph{Integrated vs. Decoupled Chunk KV Management.}

A key design decision in \paper{} is to integrate chunk-level KV reuse directly
into vLLM’s paged KV management, rather than managing chunk KV in a separate
memory pool or as an external cache. This choice reflects a fundamental
trade-off between architectural separation and system-level coherence.

A decoupled design, in which chunk KV is maintained outside the paged KV store,
appears attractive for modularity. However, such separation introduces
significant system-level costs. Because vLLM’s attention kernels are tightly
coupled to the paged KV layout, an external chunk cache must either (1) modify
the attention kernels to directly consume non-paged KV layouts—incurring
long-term maintenance burden and tight coupling to the serving engine—or (2)
copy chunk KV into the paged KV store prior to execution, introducing additional
memory traffic and latency on the critical path. In both cases, the separation
undermines the intended benefits of reuse by reintroducing data movement or
kernel fragmentation.

Integrating chunk KV reuse into the paged KV abstraction avoids these pitfalls.
By treating chunk KV as a first-class resident within the same paged block space
as prefix KV, \paper{} preserves compatibility with existing attention
execution while enabling deterministic placement, block-level sharing, and
coherent eviction. This integration allows chunk KV to participate naturally in
HBM residency management, rather than competing with prefix KV through ad hoc
buffering or copying.

This design choice is not without cost. Integration requires the scheduler to
become chunk-aware and to coordinate allocation, reuse, and eviction across both
prefix and chunk KV within a shared block pool. It also introduces resource
contention that must be managed carefully to prevent prefix-heavy workloads
from starving reusable chunks. \paper{} addresses these challenges through
explicit segmentation, canonical alignment, and reference-count–aware eviction,
but these mechanisms increase scheduler complexity relative to a purely prefix-
oriented cache.

Despite this added complexity, integration is essential for resolving several
fundamental tensions in chunk reuse. Only within the paged KV abstraction can
block alignment, selective recomputation granularity, and positional
independence be enforced coherently. Externalizing chunk KV management would
fragment these concerns across components, making it difficult to guarantee
consistent reuse semantics under multi-tenant load. Our results show that the
integrated approach yields a more stable and scalable system substrate for
practical chunk-level KV reuse.

\paragraph{Selective recomputation vs. full recomputation.}
By recomputing only the first block of each chunk, our system preserves a canonical KV representation that is consistent across requests. Compared to EPIC and CacheBlend, which recompute larger fractions of tokens, this approach significantly reduces HBM usage and redundant computation. The trade-off is that minor cross-block attention effects from omitted tokens are ignored, but empirical results show no measurable drop in accuracy for our benchmark datasets. This demonstrates that minimal selective recomputation can achieve both memory efficiency and high fidelity. In practice, recomputing fewer tokens risks missing boundary-dependent attention
effects, while recomputing more rapidly erodes the memory and latency benefits
of reuse, making single-block recomputation a practical balance point.

\paragraph{Balancing Chunk and Prefix KV Residency.}
A fundamental tension in LLM serving systems arises from the coexistence of
prefix KV and chunk KV, which differ significantly in access patterns and
lifetime characteristics. Prefix KV is typically short-lived and latency
critical, while chunk KV tends to be larger, longer-lived, and reused across
requests. Favoring either class exclusively leads to inefficiencies: prefix-
only caching overconsumes HBM due to frequent recomputation, whereas chunk-only
caching risks evicting latency-sensitive prefix state.

\paper{} addresses this tension by jointly managing prefix and chunk KV within a
shared HBM block pool, allowing allocation decisions to reflect relative reuse
and access frequency rather than fixed cache partitioning. This integrated
approach prevents either class from dominating memory resources and enables the
system to adapt to workload shifts, particularly under long-context or mixed
traffic. Our experiments with extended prompts show that maintaining this
balance is essential for avoiding HBM saturation while sustaining low latency
and stable throughput.

\paragraph{HBM residency vs. CPU/disk transfers.}
Maintaining all reusable chunks directly in HBM avoids repeated data transfers from CPU or disk, which are costly in both latency and memory bandwidth. While reserving HBM for chunk KV reduces the available memory for other tasks, our garbage-collection policies and block-alignment strategy mitigate this overhead. Proactive eviction of cold or infrequently accessed chunks reduces peak HBM usage without impacting latency, demonstrating that controlled memory management is essential for scalable deployment. This trade-off is most favorable in workloads with moderate to high chunk reuse; when reuse is sparse, retaining large chunks in HBM provides diminishing returns.

\paragraph{Position-independent KV vs. preapplied positional encodings.}
Storing KV vectors without preapplied RoPE allows chunks to be reused at arbitrary positions, but introduces the need to compute positional encodings on-the-fly during attention. This slightly increases per-token computation in the attention operator but eliminates the need for multiple per-position KV copies. Our results show that this trade-off is favorable: the runtime overhead is negligible compared to the savings in HBM and overall reduced recomputation.

\paragraph{Scalability with context length and concurrency.}
As context length grows, naive caching strategies fail to maintain efficiency due
to misaligned KV blocks and redundant recomputation; these effects are
exacerbated under concurrent, multi-tenant load. Our block-aligned, page-oriented design ensures that identical logical chunks map to the same HBM pages regardless of context offset, allowing multiple concurrent requests to share KV efficiently. This design also enables predictable scaling: throughput remains stable up to the point of HBM saturation, and TTFT remains low even for very long sessions.

Overall, these trade-offs illustrate the key design principles for production-scale LLM inference: minimal selective recomputation, semantic chunking, position-independent KV, and intelligent memory management. Together, they enable low-latency, memory-efficient, and highly scalable inference for multi-turn, long-context agent workloads.

\section{Related Work}

vLLM~\cite{2023vllm} and LMCache~\cite{cheng2025lmcache} represent the state of the art in production LLM serving. vLLM introduces paged attention and prefix caching to manage KV blocks efficiently in HBM and to reuse KV for shared prefixes across requests. LMCache lifts KV caches to a first‑class data plane across engines and tiers, exposing a connector abstraction and control APIs for pinning, lookup, movement, and compression over GPU/CPU/storage, and enabling prefix reuse and prefill–decode disaggregation at cluster scale. Recent work such as SortingHat~\cite{sortinghat2025} addresses latency-sensitive multi-GPU LLM inference by generating optimized schedules for model partitioning and execution, complementing KV-focused approaches by reducing end-to-end latency across devices. MEPIC is implemented by extending this stack: it adds an in‑HBM chunk cache coordinator inside vLLM’s paged KV allocator and plugs into LMCache’s connector and persistence layer, so that chunk‑level PIC can be deployed in existing production systems without changing the model or rewriting the serving framework.

\paragraph{Multi‑tier prefix cache scheduling.} A number of systems focus on multi-tier management and scheduling of prefix KV caches. Continuum~\cite{li2025continuum} co-designs tool-aware KV time-to-live policies with program-level scheduling to retain multi-turn agent prefixes in GPU memory when advantageous, reducing job completion time under long pauses and offloading scenarios. MCaM~\cite{chu2025mcam} extends this direction with a multi-tier KV cache spanning HBM and DRAM, using scheduler-aware pinning, pipeline prefetching, and asynchronous offloading to hide reload latency and improve end-to-end efficiency. RAGCache~\cite{jin2025ragcache} specializes multi‑level caching for RAG by organizing retrieved‑knowledge activations into a knowledge tree and caching them across GPU and host memory, while disk‑based shared KV systems such as Shared RAG‑DCache \cite{lee2025disk} enable multi‑instance sharing via a shared disk KV store. Recent work on hybrid memory architectures such as HMComp~\cite{hmcomp2025} and MEMPLEX~\cite{memplex2025} further demonstrates how careful management of HBM and DRAM (or NUMA memory nodes) can reduce swap traffic, improve locality, and accelerate memory-bound workloads. These ideas—tiered placement, prefetching, eviction policies, and metadata-aware memory allocation—can be applied on top of MEPIC’s chunk cache to manage when chunk KV should stay in HBM versus be offloaded to LMCache’s CPU/disk tiers.


\paragraph{Position-Independent Caching.} Several works have explored position-independent caching~\cite{yao2025cacheblend, yang2025kvshare, zhou20253, yang2025cacheclip, hu2024epic, yang2025kvlink}, as discussed in Section~\ref{pic}, with the common goal of reducing KV computations while maintaining accuracy by selectively recomputing only a subset of tokens. Building on this idea, Cache-Craft~\cite{agarwal2025cache} introduces a hybrid approach that adapts recomputation based on each chunk’s contextual dependencies: if a chunk is self-contained, its cache can be fully reused even in a new surrounding context; if it heavily depends on external context, full recomputation is performed; and if only a few tokens are context-dependent, those tokens are recomputed while the remaining KV states are reused. However, all of these approaches still do not optimize HBM memory usage.

\paragraph{Other context‑reusing schemes in different dimensions.} Several works attack context reuse along dimensions orthogonal to MEPIC. GenCache~\cite{chakraborty2025generative} operates at the plaintext level: it caches and synthesizes responses for structurally similar prompts, providing a parallel solution that does not manipulate KV caches at all. RAGBOOST~\cite{jiang2025ragboost} increases prefix cache hit rates by reordering chunk order and positions across concurrent sessions and multi‑turn interactions, maximizing shared prefixes where prompt reordering is semantically safe; in contrast, MEPIC assumes prompts are fixed and instead changes how chunk KV is represented and shared in HBM. HeteroRAG~\cite{liu2025heterrag} explores heterogeneous processing‑in‑memory (PIM) architectures for RAG, combining HBM‑ and DIMM‑based PIM and leveraging CacheBlend‑style ideas, but it requires specialized hardware and does not spell out how to integrate cross‑request chunk reuse into a software paged‑KV engine; MEPIC is purely a software mechanism and can coexist with such hardware accelerators.

\paragraph{Operator‑level acceleration.} FlashForge~\cite{wang2025flashforge} and MoSKA~\cite{rhee2025moska} operate at the attention‑kernel level rather than in the KV cache manager. FlashForge proposes a shared‑prefix attention kernel that combines memory accesses for shared prefixes across requests to reduce decode‑time memory traffic and accelerate per‑token latency. MoSKA introduces Shared KV Attention to turn attention over shared context from a series of memory‑bound GEMV operations into a single compute‑bound GEMM, aided by sparse attention and disaggregated infrastructure. Multi-core AI accelerators with managed caches~\cite{aiaccel2025} and GPU memory systems such as DREAM~\cite{dream2025} further optimize memory traffic and latency at the hardware level. These techniques are complementary to MEPIC: MEPIC provides page‑aligned, in‑HBM chunk reuse at the KV layer, while FlashForge/MoSKA can be layered underneath to further accelerate attention computation over the shared prefixes and chunks that MEPIC keeps resident in HBM.

\paragraph{KV Compression.}
A complementary line of work reduces the memory footprint of the KV cache by compressing KV states during generation. A prominent class of approaches focuses on \emph{KV cache quantization}, which lowers per-entry memory cost by storing keys and values at reduced precision. Recent work demonstrates that aggressive and adaptive quantization of KV caches can substantially reduce memory usage with minimal impact on generation quality or latency \cite{hooper2024kvquant, cheng2025qaq, he2024zipcache}.

Beyond quantization, several methods exploit structural properties of attention to enable more aggressive compression. FastGen and RazorAttention leverage head-level heterogeneity, selectively retaining or discarding KV states based on attention locality or the presence of retrieval-oriented heads, achieving significant memory savings without retraining \cite{ge2024fastgen, tang2024razorattention}. Other approaches target orthogonal dimensions of redundancy: MiniCache compresses KV caches across layers by merging similar states in depth, while CacheGen focuses on compressing KV caches for efficient transmission and fast context loading in distributed serving systems \cite{liu2024minicache, liu2024cachegen}.

Our approach is orthogonal to these KV compression techniques. Rather than modifying the precision, structure, or lifetime of KV entries, \paper{} improves inference efficiency by enabling deterministic reuse of KV caches across requests through page-aligned KV placement. As a result, \paper{} can be seamlessly combined with existing KV compression methods to jointly improve inference speed, memory efficiency, and serving scalability.

\section{Conclusion \& Future Work}
In this work, we presented a system for efficient chunk-level KV reuse in production-scale LLM serving. We identified key limitations in existing prefix caching and position-independent caching approaches, including lack of in-HBM chunk management, misalignment with paged KV storage, per-request KV divergence, and the need for model modifications. To address these challenges, we introduced a Chunk Cache Coordinator for HBM residency, a segmentation and padding scheme for page-alignment, a selective recomputation strategy to preserve canonical KV, and a positional-encoding-free KV format supporting reuse across arbitrary positions. Together, these techniques enable page-aligned, position-independent chunk reuse, improving cache hit rates, reducing recomputation, and lowering end-to-end inference cost.

Looking forward, several directions can further extend the capabilities of our work. Building on our canonical chunk KV abstraction, dynamic chunk prioritization and heat-aware eviction policies could optimize HBM utilization under high-concurrency workloads. Extending position-independent reuse to multi-modal models and cross-attention scenarios would broaden the applicability of our approach to more complex LLM interactions. Integrating the chunk-aware cache with model quantization or memory compression techniques could reduce overall memory footprint while maintaining reuse efficiency. Finally, exploring privacy-preserving shared KV caches across users or tenants could enable collaborative multi-session caching without exposing sensitive content. These directions leverage the core design principles introduced in this work, highlighting opportunities to enhance inference efficiency and scalability in production LLM serving.



\newpage
\bibliographystyle{ACM-Reference-Format}
\bibliography{sample-base}


\end{CJK*}
\end{document}